\documentclass[runningheads]{llncs}
\usepackage{microtype}
\usepackage{graphicx}
\usepackage{adjustbox}
\usepackage{caption}
\usepackage{subcaption}
\usepackage{booktabs}
\usepackage{amsmath}
\usepackage{amssymb}
\usepackage{mathtools}
\usepackage{multirow}
\begin{document}
\title{Test-Time Adaptation with Principal Component Analysis}
%
%
\author{Thomas Cordier\inst{1, 2}\orcidID{0000-0002-1314-9619} \and
Victor Bouvier\inst{3} \and
Gilles Hénaff\inst{1} \and
Céline Hudelot\inst{2}}
\authorrunning{T. Cordier et al.}
%
\institute{Thales Land and Air Systems, 2 Avenue Gay-Lussac, 78990 Elancourt, France \and
Universit\'{e} Paris-Saclay, CentraleSup\'{e}lec, Math\'{e}matiques et Informatique pour la Complexit\'{e} et les Syst\`{e}mes, 91190, Gif-sur-Yvette, France \and
Dataiku, 203 Rue de Bercy, 75012, Paris, France}
\maketitle              
\begin{abstract}
Machine Learning models are prone to fail when test data are different from training data, a situation often encountered in real applications known as distribution shift. While still valid, the training-time knowledge becomes less effective, requiring a test-time adaptation to maintain high performance. Following approaches that assume batch-norm layer and use their statistics for adaptation\cite{nado2021evaluating}, we propose a Test-Time Adaptation with Principal Component Analysis (TTAwPCA), which presumes a fitted PCA and adapts at test time a spectral filter based on the singular values of the PCA for robustness to corruptions.
TTAwPCA combines three components: the output of a given layer is decomposed using a Principal Component Analysis (PCA), filtered by a penalization of its singular values, and reconstructed with the PCA inverse transform. 
This generic enhancement adds fewer parameters than current methods\cite{mummadi2021testtime,sun19ttt,wang2021tent}. Experiments on CIFAR-10-C and CIFAR-100-C\cite{hendrycks2018benchmarking} demonstrate the effectiveness and limits of our method using a unique filter of 2000 parameters.

\keywords{Robustness  \and Test-time Adaptation \and Principal Component Analysis \and Filtering.}
\end{abstract}
\section{Introduction}
Deep neural networks are optimized to achieve high accuracy on their training data, given the hypothesis that test data will follow the same distribution during inference. However, distribution shift occurs in many industrial applications, for instance, when a sensor malfunctions. The accuracy of a predictive task drops as the test data shifts from its training distribution \cite{hendrycks2018benchmarking,Candeladatashift2009}. Domain adaptation prevents such failures by jointly training on source and target data. Instead, Test-time adaptation mitigates the domain gap either by test-time training or fully test-time adaptation according to the availability of source data. Test-time training (TTT) augments the training objective on source data with an unsupervised task that remains at test time to optimize domain-invariant representations. \textit{Fully test-time adaptation} \cite{wang2021tent} does not alter training and only needs testing observations and a pre-trained model for privacy, applicability, or profit \cite{Chidlovskii}.

 To enhance generalization, Spectral regularization \cite{Bartlett2017spectral} especially for GANs \cite{miyato2018spectral} and $L^2-$regularization are standard tools during training \cite{Neyshabur2017generalization}. $L^2-$regularization reduces model variance for different potential training sets and constrains the model complexity by lowering the weights of its layers. Spectral normalization penalizes the weight matrices by their largest singular value to ensure the Lipschitz continuity of the neural network. This property provides an upper bound for the perturbations of the outputs of the model, preventing exploding and vanishing gradients. 
 
 Taking inspiration from these previous works, we aim to learn the best fitting parameters of a spectral filter on a corrupted dataset without supervision. We introduce TTAwPCA, which projects a batch of inputs onto a spectral basis, filters the projected data points, and reconstructs the filtered batch. As \cite{wang2021tent}, we minimize entropy to learn the parameters of the filter. This generic unsupervised learning loss makes few assumptions about the data.
 
 In this paper, we first overview state-of-the-art test-time adaptation (Sec. \ref{sec2}). Then, we introduce a simple yet effective method: TTAwPCA (Sec. \ref{sec3}). We demonstrate its effectiveness experimentally in tackling corrupted data (Sec. \ref{sec4} and we discuss our results compared with other methods (Sec. \ref{sec5}).
 
 \section{Related work}
\label{sec2}
\textbf{Unsupervised Domain Adaptation} jointly adapts on source and target domain through transduction, thus requiring both simultaneously. Several properties have been optimized: cross-domain feature alignment\cite{Gretton_2009,CORAL2017,Candeladatashift2009}, adversarial invariance\cite{Tzeng_2017_CVPR,pmlr-v37-ganin15,JMLR:v17:15-239,pmlr-v80-hoffman18a}, and shared proxy tasks \cite{sun2019unsupervised} such as predicting rotation and position. In our work, we want to use only the target domain at test time. 

\noindent\textbf{Test-time adaptation} indicates methods tackling the domain gap during inference.
\textit{TTT}\cite{sun19ttt} augments the supervised training objective with a self-supervised loss using source data. Only the self-supervised loss keeps adapting at test time on target domain. It relies on predicting the rotation of inputs, a visual proxy task, but designing suitable proxy tasks can be challenging. Training parameters are altered during training and test-time adaptation.
\textit{Test-time batch normalization}\cite{NEURIPS2020_85690f81,nado2021evaluating} allows statistics of batch norm layers to be tracked during the distribution shift at test time.
Test ENTropy minimization (\textit{TENT})\cite{wang2021tent} exhibits adaptation at test time of feature modulators extracted from spatial batch normalization to circumvent the distribution shift. Entropy minimization is a generic and standard loss for domain adaptation to penalize classes overlap. 
Information maximization \cite{NIPS2010_42998cf3,ICML2012Shi_566,ICML-2017-HuMTMS} used by \cite{liang2020shot,mummadi2021testtime} involves entropy minimization and diversity regularization. The diversity regularizer averts collapsed solutions of entropy minimization. Soft Likelihood Ratio and Input Transformation (\textit{SLR+IT})\cite{mummadi2021testtime} argues that Information maximization compensates for the vanishing gradient issues of entropy minimization for high confidence predictions. Moreover, an additional trainable network shares the input samples with the tested network to partially correct the domain shift.
Principal Component Analysis cuts out noisy eigenvalues to remove uncorrelated noise\cite{pca_denoising_audio,pca_denoising_image}. In addition, we propose to add fully test-time learnable parameters to reduce the remaining noise of corrupted data onto the spectral basis.

TTAwPCA, as explained next, uses fewer parameters. Those parameters are only used for fully test-time  adaptation, facilitating the activation or the removal of the adaptation in real-time. Our method also relies on entropy minimization.

\section{Adapting a spectral filter}
\label{sec3}
In this section, we explain how Principal Component Analysis can be used to create a partial suppression of unwanted eigenvalues of a matrix, namely a spectral filter. Then, we extend this notion to adaptation of a neural network at test-time.

Principal Component Analysis (PCA) linearly separates multivariate systemic variation from noise. Consider $A$ an $N \times p$ data matrix. PCA defines its principal components as the $q \leq p$ unit vectors such that the $i$-th vector satisfy orthogonality with the first $i-1$ and best fits the direction of data. The process performs a change of basis on the data according to the principal components. They are computed by Singular Value Decomposition (SVD) of $A$ and ranked top down by the corresponding singular value scale. Noisy thus irrelevant principal components can be ignored. This can be seen as a high-pass filter on the principal components. More technically, incremental PCA can be performed if the dataset is too large to fit in the memory. Incremental PCA uses an amount of memory independent of the number of input data samples to build a low-rank approximation. 

Let a neural network $f_{\theta}$ with parameters $\theta$ be trained to completion on a source set $X_{\mathcal{D}}$ of $N$ samples from a distribution $\mathcal{D}$. Parameters $\theta$ are thus frozen after training. The initialization of our method takes place before testing. TTAwPCA is added after the $j$th layer. It consists of a Principal Component Analysis (PCA) and, for now, a pass-through filter. To fit its PCA, the concatenated output $A_{j, \mathcal{D}}$ of the $j$th layer has to be flattened from the shape $N$ elements of the batch times $c$ channels times the spatial dimensions $h \times w$ to a rectangular matrix of size $N \times p$ where $p = c \cdot h \cdot w$ and then mean normalized. Singular Value Decomposition breaks down the flattened training output $\tilde{A_{\mathcal{D}}}$ as:
\begin{equation}
   {A_{j,\mathcal{D}}} = U \Lambda V^\top 
\end{equation}
where $\Lambda$ is an $N \times p$ matrix of singular values, $U$ an $N \times N$ matrix of left singular vectors and $V$ an $p \times p$ matrix of right singular vectors. We define a hyperparameter $L$ such that only the first $L$ singular values are conserved. Note that this operation belongs to the training procedure.

At test time, the filter $F_\Gamma$ is enabled to optimize its parameters $\Gamma=\{\gamma_i; i\in[0,L-1]\}$ of the corrupted singular values. Let the $t$-th batch of corrupted observation $x_t \sim D'$ be presented to the model $f_{\theta, \Gamma}$. Let ${A_{j, \mathcal{D'}, t}}$ be the $t$th batched output of the $j$th layer. After the flatten operation and the mean normalization, ${A_{j, \mathcal{D'}, t}}$ is projected onto the singular basis vectors by $V_L$, filtered by $F_\Gamma$ and reconstructed by $V_L^\top$ as $O_{t, \mathcal{D'}}$ in its original basis:
\begin{equation}
   O_{t, \mathcal{D'}} = {A_{j,\mathcal{D'}, t}} V_L F_\Gamma V_L^\top 
\end{equation}
We designed a filter $F_\Gamma$ related with $L^2-$regularization as demonstrated in \ref{L2Regul} of diagonal element $F_{i,i}$ based on the singular values $\Lambda_{L}=\{\lambda_i; i\in[0,L-1]\}$ of the training set and $L$ learning parameters $\gamma_i$:
\begin{equation}
   F_{i,i}(\gamma_i) = \frac{\lambda_{i}}{\lambda_{i} + \mathrm{ReLU}(\gamma_{i})}
\end{equation}
The $\mathrm{ReLU}$ activation assures the positivity of the filter, increasing its stability.

Similarly, we designed a negative exponential filter $F_\Gamma$ of diagonal element $F_{i,i}$:
\begin{equation}
\label{intro_filter}
   F_{i,i}(\gamma_i) = \frac{1}{1+\exp(\gamma_i^2-\lambda_i)}
\end{equation}

We denote this model $f_{\theta, \Gamma}$ composed of $f_\theta$ and TTAwPCA. The learning parameters $\Gamma$ are optimised over the batch $x_t$ using entropy minimization of model prediction $\hat{y}_t = f_{\theta, \Gamma}(x_t)$ as test-time objective:
\begin{equation}
   L(x_t) = H(\hat{y}_t)=-\sum_c p(\hat{y}_{t, c})\log p(\hat{y}_{t, c})
\end{equation}

\section{Experiments}
\label{sec4}
\begin{table*}[t]
\begin{center}
\begin{footnotesize}
\caption{Episodic classification error benchmark on CIFAR-10-C and CIFAR-100-C with the highest severity [in \%].}
\label{tab:offlinecorruption}

\begin{adjustbox}{width=\textwidth}

\begin{tabular}{cccccccccccccccccc}
\toprule
Dataset & Method & Mean & Gauss &  Shot &  Impulse &  Defocus &  Glass &  Motion &  Zoom &  Snow &  Frost &   Fog &  Bright &  Contrast &  Elastic &  Pixel &  JPEG \\
     &        &        &       &          &          &        &         &       &       &        &       &         &           &          &        &       &       \\
\midrule
\multirow{5}{*}{CIFAR-10-C} 
& No Adaptation &   43.53 &   72.33 &  65.71 &     72.92 &    46.94 &   54.32 &   34.3 & 42.02 & 25.07 &  41.30 & 26.01 &   9.30 &      46.69 &    26.59 &  58.45 & 30.30 \\
& BN    &  20.44 &   28.08 & 26.12 &    36.27 &    12.82 &  35.28 &   14.17 & \textbf{12.13} & 17.28 &  17.39 & 15.26 &   8.39 &     12.63 &    23.76 &  19.66 & 27.30 \\
& TENT   &  \textbf{19.96} &   28.05 & 26.11 &    36.31 &    12.80 &  35.28 &   14.16 & 12.14 & \textbf{17.27} &  \textbf{17.36} & 15.23 &   8.37 &     \textbf{12.59} &    23.77 &  \textbf{19.61} & 27.31 \\
& exp-TTAwPCA (ours)     &  20.35 &   \textbf{25.5} & \textbf{23.55} &    \textbf{33.77} &    14.82 &  35.04 &   15.24 & 13.76 & 17.73 & 17.43 &  16.09 & 8.62 &   14.58 &     24.44 &    20.00 &  \textbf{24.68} \\
& ReLU-TTAwPCA (ours) &   20.42 &   28.10 &  25.99 &     36.13 &    \textbf{12.72} &   \textbf{34.93} &   \textbf{14.00} & 12.24 & 17.29 &  17.8 & \textbf{15.07} &   \textbf{8.26} &      13.09 &    \textbf{23.47} & 19.76 & 27.41 \\
\midrule
\multirow{5}{*}{CIFAR-100-C} 
& No Adaptation &   85.54 &   93.84 &  93.60 &     96.63 &    91.49 &   92.79 &   86.51 & 88.69 & 70.91 &  82.30 & 84.74 &   47.26 &      96.30 &    85.02 &  89.50 & 83.49 \\
& BN    &  36.61 &   47.21 & 46.72 &    55.59 &    \textbf{27.33} &  47.75 &   \textbf{28.23} & 26.65 & 32.74 & 33.63 & 32.92 &   \textbf{21.35} &      29.64 &    {37.79} &  33.99 & 47.56 \\
& TENT   &  \textbf{34.56} &   \textbf{42.91} & \textbf{41.94} &    \textbf{49.76} &    {28.27} &  \textbf{44.55} &   {28.75} & {27.38} & \textbf{30.99} &  \textbf{31.59} & \textbf{30.72} &   {21.88} &    {30.81} &    \textbf{35.42} &  \textbf{31.27} & \textbf{42.09} \\
& exp-TTAwPCA (ours)     &  37.89 &   45.92 & 45.71 &    {54.23} &    32.82 &  {47.88} &   31.98 & 30.04 & 33.53 & 35.12 &  36.26 & 22.46 &   32.92 &     39.18 &    34.91 &  {45.37} \\
& ReLU-TTAwPCA (ours) &   36.62 &   47.41 &  46.80 &     55.50 &    27.61 &   47.76 &   28.28 & \textbf{26.54} & 32.67 &  33.46 & 32.80 &   21.41 &      \textbf{29.55} &    37.67 & 34.25 & 47.53 \\

\bottomrule
\end{tabular}

\end{adjustbox}
\end{footnotesize}
\end{center}
\end{table*}

\begin{table*}[t]
\begin{center}
\begin{footnotesize}
\caption{Online classification error benchmark on CIFAR-10-C and CIFAR-100-C with the highest severity [in \%].}
\label{tab:onlinecorruption}

\begin{adjustbox}{width=\textwidth}

\begin{tabular}{cccccccccccccccccc}
\toprule
Dataset & Method & Mean & Gauss &  Shot &  Impulse &  Defocus &  Glass &  Motion &  Zoom &  Snow &  Frost &   Fog &  Bright &  Contrast &  Elastic &  Pixel &  JPEG \\
     &        &        &       &          &          &        &         &       &       &        &       &         &           &          &        &       &       \\
\midrule
\multirow{3}{*}{CIFAR-10-C} 
& TENT   &  \textbf{18.57} &   \textbf{25.09} & \textbf{22.76} &    \textbf{32.71} &    \textbf{12.01} &  \textbf{31.88} &   \textbf{13.25} & \textbf{11.12} & \textbf{15.9} &  \textbf{16.32} $\pm$ 0.59 & \textbf{13.82} &   \textbf{8.21} &     \textbf{11.66} &    \textbf{22.02} &  \textbf{17.29} & \textbf{24.5} $\pm$ 0.43 \\
& exp-TTAwPCA (ours)     &  20.28 &   25.42 & 23.44 &    33.92 &    14.79 &  34.81 &   15.18 & 13.71 & 17.52 & 17.53 $\pm$ 0.17 &  16.09 & 8.62 &   14.58 &     24.44 &    20.00 &  \textbf{24.68} $\pm$ \textbf{0.12} \\
& ReLU-TTAwPCA (ours) &   20.45 &   28.14 &  25.84 &     36.23 &    12.85 &   35.04 &   14.01 & 12.22 & 17.27 &  17.63 & 15.08 &   8.37 &      13.05 &    23.58 & 19.93 & 27.44 \\
\midrule
\multirow{3}{*}{CIFAR-100-C} 
& TENT   &  \textbf{31.7} &   \textbf{38.74} & \textbf{36.88} &    \textbf{44.00} &    \textbf{26.91} &  \textbf{41.03} &   \textbf{27.33} & \textbf{25.54} & \textbf{28.18} &  \textbf{28.85} & \textbf{28.03} &   \textbf{20.44} &     \textbf{28.81} &    \textbf{33.93} &  \textbf{28.41} & \textbf{38.41} \\
& exp-TTAwPCA (ours)     &  37.89 &   {46.02} & {45.8} &    {54.15} &    32.56 &  {47.87} &  31.91 & 30.14 & 33.62 & 35.19 &  35.98 & 22.33 &   33.08 &     39.18 &    34.93 &  {45.51} \\
& ReLU-TTAwPCA (ours) &   36.83 &   47.39 &  46.82 &     55.95 &    27.80 &   48.30 &   28.49 & 26.85 & 32.92 &  33.78$\pm$ 0.20 & 32.91 &   21.64 &      29.58 &    37.94 & 34.48 & 47.59$\pm$ 0.10 \\

\bottomrule
\end{tabular}

\end{adjustbox}
\end{footnotesize}
\end{center}
\end{table*}

In this section, we evaluate TTAwPCA on a classification benchmark with models trained on CIFAR-10 or CIFAR-100 \cite{CIFAR10} and evaluated on CIFAR-10-C or CIFAR-100-C \cite{hendrycks2018benchmarking} respectively. We describe the experimental setup in the next section followed by ablations on the hyperparameters of our method.

\subsection{Experimental Setup}
\noindent\textbf{Dataset}. CIFAR-10-C and CIFAR-100-C \cite{hendrycks2018benchmarking} are standard image classification datasets for domain adaptation issues. Both datasets are build on 15 copies of the test sets of CIFAR-10 and CIFAR-100 \cite{CIFAR10} respectively each augmented by a corruption with five severity levels. As the only difference between uncorrupted and corrupted datasets lies in the specific corruption given its severity, the stability of the neural network is directly explained by its loss of performance.

\noindent\textbf{Models}. We use the publicly available pre-trained WideResNet-28-10 \cite{Zagoruyko2016WRN} of RobustBench \cite{croce2021robustbench}. We trained a model on CIFAR-100 achieving 83\% accuracy on the test set, as Robustbench does not provide one. TTAwPCA is set after the first convolutional layer with only 2000 parameters for our best results on both datasets.
We compare our two different filters with TENT \cite{wang2021tent} and test-time batch statistics updates \cite{NEURIPS2020_85690f81,nado2021evaluating}.

\noindent\textbf{Settings}. The evaluation is conducted as following: the model is loaded accordingly with the corrupted test set it will perform on. For each corruption at a given severity, the model adapts itself depending on the episodic or online adaptation. Episodic and online settings describe whether the model is reset after optimization on each batch or after optimization on the corruption at a given severity. Episodic test-time adaptation only optimizes the model on a single batch, then evaluates the classification error on this batch and finally resets the model before receiving a new batch until the test set has been tested. Online test-time adaptation optimizes over the test set, batch after batch, without forgetting. Both settings have their reasons: online TTA can mitigate the corruption on a wider set of experiments but it can also fail into catastrophic solutions as corruption remains unknown and can be unrealistic. On the other hand, episodic TTA seems more stable to degraded solution as the model is reset after each batch optimization.

\noindent\textbf{Optimization}. We optimize the parameters $\Gamma$ of the filter by Adam \cite{DBLP:journals/corr/KingmaB14} for one step on both offline and episodic fully test-time adaptation settings. We set the batch size at 200 samples and the learning rate at 0,001. $L=2000$ proved to be sufficient for our method, as shown in the next section.

\noindent\textbf{Metric}. To evaluate the ability of the model to still classify correctly the corrupted data of CIFAR-10-C and CIFAR-100-C according to the classes of CIFAR-10 and CIFAR-100, we measure the classification error. Each corruption at each severity are taken as different test sets. We also adjoin the average over each corruption at a given severity for more clarity.

\subsection{Ablation Studies}
\subsubsection{PCA rank and parameters of the filter}
\label{AS_rank}
\begin{figure}[htbp]
\centering
\begin{subfigure}{0.5\textwidth}
  \centering
  \includegraphics[width=\linewidth]{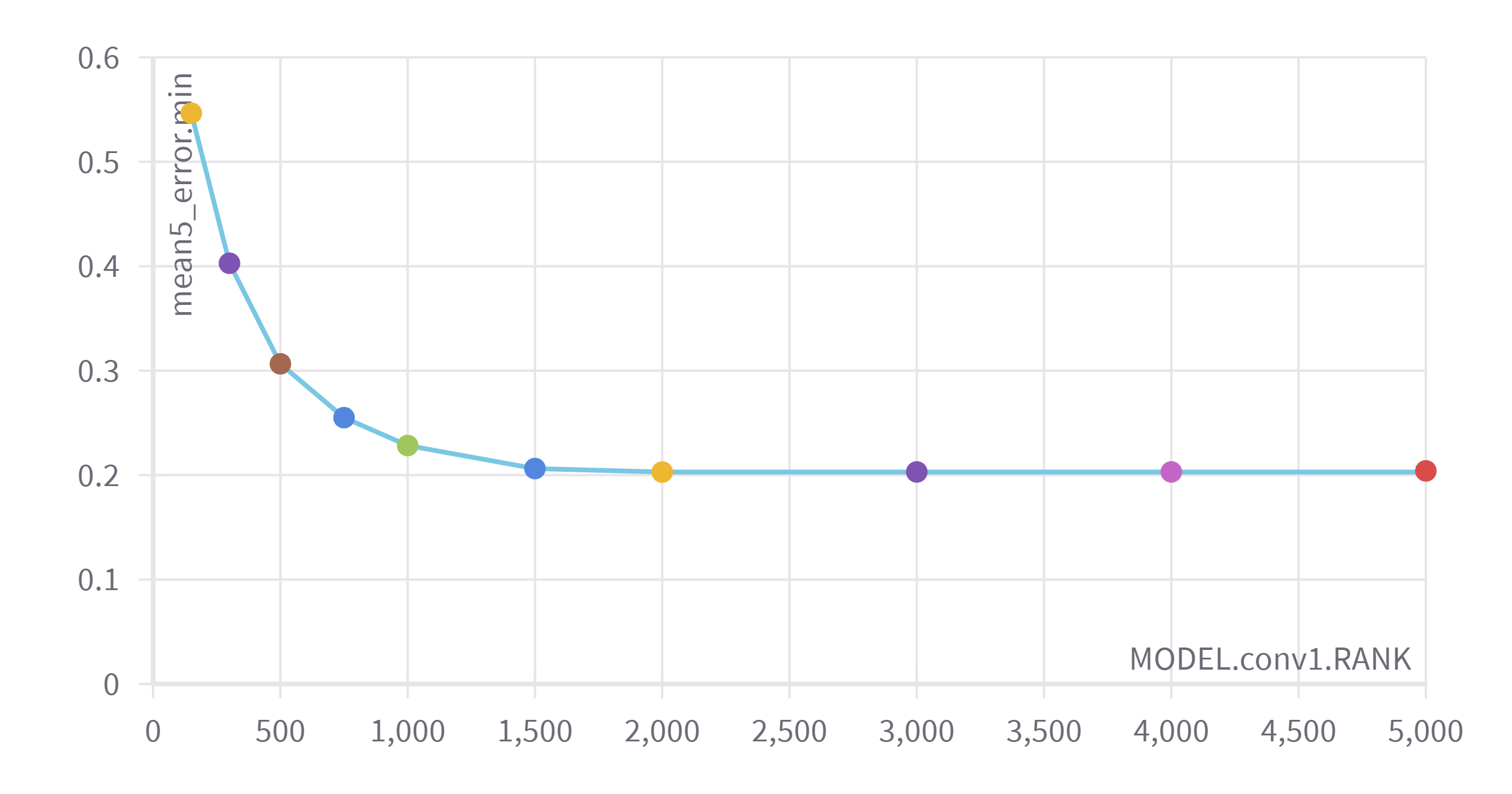}
  \caption{CIFAR-10-C}
  \label{AS_rank:sub1}
\end{subfigure}%
\begin{subfigure}{0.5\textwidth}
  \centering
  \includegraphics[width=\linewidth]{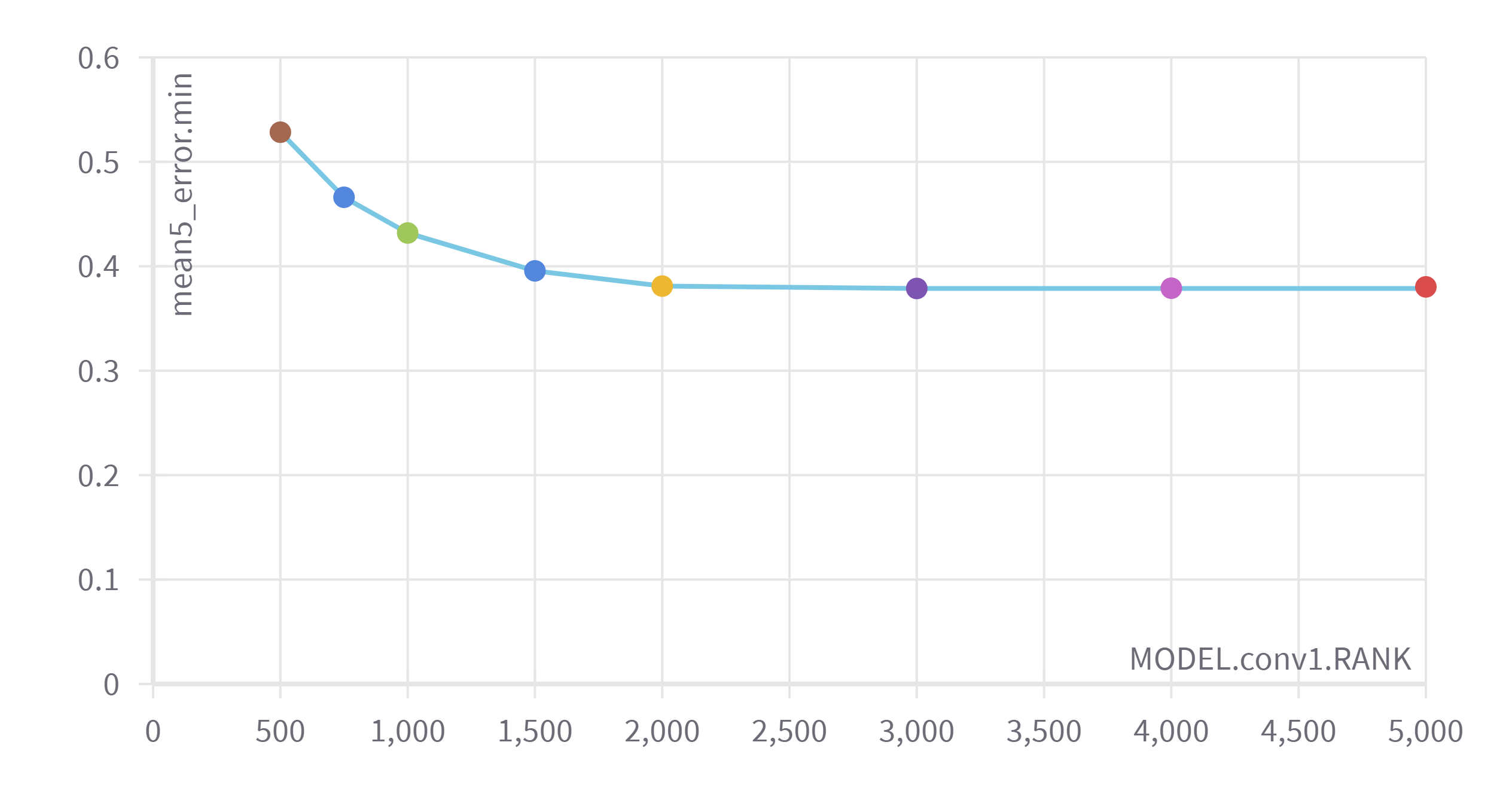}
  \caption{CIFAR-100-C}
  \label{AS_rank:sub2}
\end{subfigure}

\caption{Episodic mean error along all corruptions at severity 5 for different rank of the PCA of TTAwPCA.}
\label{fig:rank}
\end{figure}

Our experiments investigated how many parameters are enough to tackle corrupted data points. While these results only apply to CIFAR-10-C and CIFAR-100-C, we experienced that 2000 parameters are enough to effectively train a model to regain accuracy after a distributional shift at test time. In Figure \ref{fig:rank}, we show the mean error on all corruptions at severity 5 for different ranks of the PCA on both datasets. We averaged over three runs for each PCA rank with minor variations. The optimization has been done in an episodic setting.

\subsubsection{Optimizing steps}
\label{AS_steps}
\begin{figure}[htbp]
\centering
\begin{subfigure}{0.5\textwidth}
  \centering
  \includegraphics[width=\linewidth]{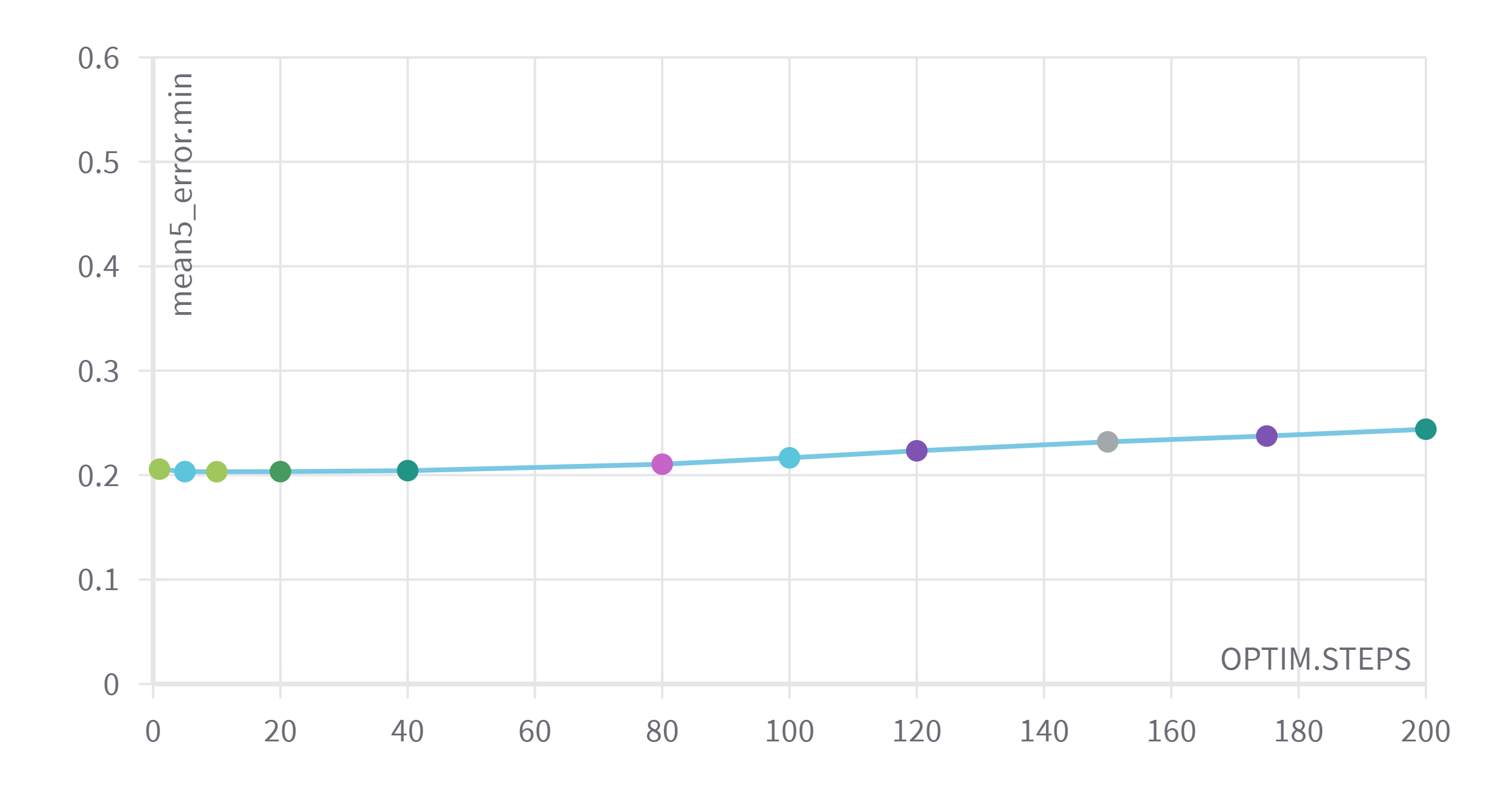}
  \caption{CIFAR-10-C}
  \label{AS_steps:sub1}
\end{subfigure}%
\begin{subfigure}{0.5\textwidth}
  \centering
  \includegraphics[width=\linewidth]{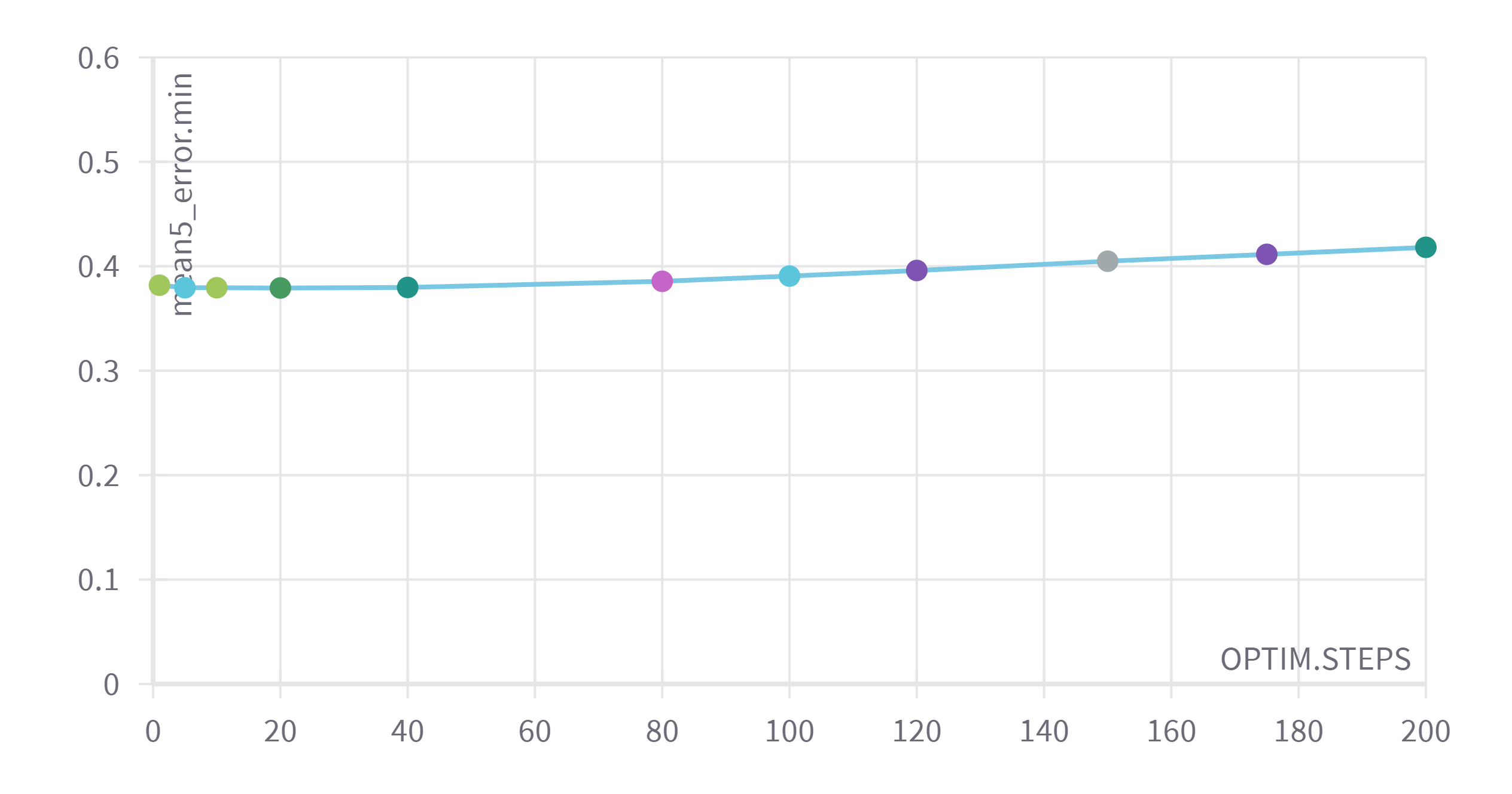}
  \caption{CIFAR-100-C}
  \label{AS_steps:sub2}
\end{subfigure}

\caption{Online mean error along all corruptions at severity 5 for different number of learning steps of TTAwPCA.}
\label{fig:steps}
\end{figure}

As shown in \cite{mummadi2021testtime}, error degrades over optimization steps as entropy minimization lacks target distribution regularization. Still, this effect is minor compared with the accuracy retrieval achieved by our simple method.


\section{Discussion}
\label{sec5}
TTAwPCA tackles common corruptions \cite{hendrycks2018benchmarking} by improving the accuracy of each perturbed set. With only the 2000 parameters, TTAwPCA achieves state-of-the-art performance on various corruptions in the episodic CIFAR-10-C setting. Namely: Gaussian Noise, Shot Noise, Impulse Noise, Glass Blur, and JPEG compression for the exponential filter and Defocus Blur, Glass Blur, Motion Blur, Fog, Brightness, and Elastic Transformation for the ReLU filter whereas performing close to TENT \cite{wang2021tent} on the rest. For the record, TENT needs 17952 parameters to adapt a WRN-28-10 which is close to nine times more. So our method achieves a better trade-off between accuracy retrieval and the number of parameters. On the other hand, TTAwPCA does not take advantage of the online setting and does not scale well to CIFAR-100-C. We provide intuitions to explain this observation.



TTAwPCA enables PCA to filter noisy singular values on the remaining dimensions, assuming additive noises increase singular values. However, we observe some corruptions to reduce singular values effectively, thus filtering crucial information to the tested task. A penalizing filter is unable to recover this loss of information. Adding a multiplicative parameter to each diagonal element of our filter became a subject of our interest but was found unstable. To increase stability, we normalized each singular value $\lambda_i$ by its higher value: $\lambda_0$. 
The instability of the tested filter prevents its convergence in an online setting.

Our results on CIFAR-100-C tend to be underperforming. High similarity between classes of CIFAR-100 might be too complex for TTAwPCA to reach over-parametrized methods such as TENT. A subtle change in the first principal components of the PCA can significantly affect the discriminability of the model if corruption occurs and the classes are too close. The first convolutional layer might not be discriminative enough to perform reliable principal components. On the other hand, the following layers merge the corruption and the features relevant to the task.

We argue that TTAwPCA follows the setting of \textit{Fully test-time adaptation} \cite{wang2021tent} as TTAwPCA does not change the training objective. TTAwPCA expects a model to have a fitted PCA after completing the training procedure. Equivalently TENT needs spatial batch normalization layers to operate.

Lastly, TTAwPCA is the only method that does not alter any training parameter. Its test-time update can be fully deactivated without reloading the model instead of TENT or batch adaptation at test time (BN). The batch normalization parameters are forgotten through their processes. PCA also offers a linear adaptation of the model.


\section{Conclusion}
This paper introduced a new layer called TTAwPCA, filtering the singular values to tackle the out-of-distribution shift at test time. This spectral filter, initialized after training, is optimized on the test dataset with a task agnostic loss. We compared the effectiveness of our method in an online and an episodic setting to TENT \cite{wang2021tent} on CIFAR-10-C and CIFAR-100-C \cite{hendrycks2018benchmarking}. We argue our technique to adapt efficiently, reaching a new state-of-the-art on some corruptions without altering training parameters. We provided explanations of the success and the flaws of spectral penalization and its connections with standard methods in Machine Learning.
\newpage

\newpage
\appendix
\onecolumn
\section{Appendix}
\subsection{Connection with $L^2-$Regularization}
\label{L2Regul}
Let $X \in \mathbb R^{n\times d}$, where $n$ is the number of samples and $d$ is the number of features. We consider the simple case of linear regression where $Y = X \theta$ where $\theta$ is the parameter of the model. The optimal parameter are defined as follows:
\begin{equation}
    \theta^\star := \arg \min_\theta || Y - X \theta||^2
\end{equation}
and it is straightforward to observe the following closed form: 
\begin{equation}
    \theta^\star = \left (X^\top X \right )^{-1} X^\top Y 
\end{equation}
it is also straightforward to observe that: 
\begin{equation}
    \theta^\star_\gamma := \arg \min_\theta || Y - X \theta||^2 + \gamma \cdot || \theta||^2 
\end{equation}
leads to the close form: 
\begin{equation}
    \theta^\star_\gamma = \left (X^\top X+ \gamma I_d  \right )^{-1} X^\top Y 
\end{equation}

In the following, we note $C = X^\top X$. $C$ is has an orthogonal eigen decomposition (symmetric, positive and definite). 
\begin{equation}
    C = U^\top D U
\end{equation}
where $U \in \mathbb U(d)$ which is the unitary group $U^\top U = I_d$.  We note the basis change of $X$ as follows: 
\begin{equation}
    \tilde X := X U^\top 
\end{equation}
By construction, $\tilde X$ has a diagonal covariance, 
\begin{equation}
    \tilde X^\top X = U X^\top X U^\top = U X^\top X U^\top = U C U^\top = D
\end{equation}
Now, what happens when regressing from $\tilde X$ to obtain $\tilde \theta^\star$:
\begin{equation}
    \tilde \theta^\star := D^{-1} \tilde X Y 
\end{equation}
Now,
\begin{equation}
   \tilde X \tilde \theta^\star = \tilde X D^{-1} \tilde X^\top Y = Y 
\end{equation}

\begin{equation}
   X \underbrace{U^\top  D^{-1} U X^\top  Y}_{=\theta}   = \tilde X D^{-1} \tilde X^\top Y = Y 
\end{equation}

\begin{equation}
   X \underbrace{U^\top  (D+\gamma I_d) ^{-1} U X^\top  Y}_{=\theta}   = \tilde X D^{-1} \tilde X^\top Y = Y 
\end{equation}

\begin{equation}
    \theta^\star \tilde X = \theta^\star X 
\end{equation}

Let break the equation of $\theta_\gamma^\star$:

\begin{align}
     \theta^\star_\gamma &= \left (X^\top X+ \gamma I_d  \right )^{-1} X^\top Y \\
     & =  \left ( U^\top (D+ \gamma I_d)U  \right )^{-1} X^\top Y \\
     & =   U^\top (D+ \gamma I_d)^{-1}U   X^\top Y \\
     & =   U^\top \underbrace{D (D+ \gamma I_d)^{-1}}_{F_\gamma} D^{-1} U   X^\top Y \\
    & =   U^\top F_\gamma U U^\top D^{-1} U   X^\top Y \\
    & = U^\top F_\gamma U \theta^\star_0
\end{align}
where $F_\gamma$ is a diagonal matrix such that:
\begin{equation}
    F_{\gamma, i, i} = \frac{\lambda_i}{\lambda_i + \gamma}
\end{equation}
where $\lambda_i$ is the $i-$th eigen-value of $C$.

\subsection{Insight on CIFAR-10-C}
\begin{figure*}[hbt!]
   \centering
\begin{tabular}{ccccc}
Gaussian Noise & Shot Noise & Impulse Noise & Defocus Blur & Glass Blur\\
\includegraphics[width=1cm, height=1cm, keepaspectratio]{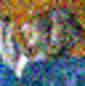}&
\includegraphics[width=1cm, height=1cm, keepaspectratio]{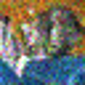}&
\includegraphics[width=1cm, height=1cm, keepaspectratio]{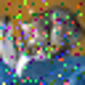}&
\includegraphics[width=1cm, height=1cm, keepaspectratio]{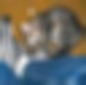}&
\includegraphics[width=1cm, height=1cm, keepaspectratio]{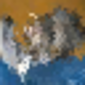}\\
Motion Blur & Zoom Blur & Snow & Frost & Fog\\
\includegraphics[width=1cm, height=1cm, keepaspectratio]{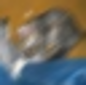}&
\includegraphics[width=1cm, height=1cm, keepaspectratio]{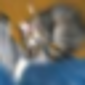}&
\includegraphics[width=1cm, height=1cm, keepaspectratio]{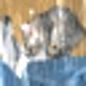}&
\includegraphics[width=1cm, height=1cm, keepaspectratio]{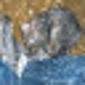}&
\includegraphics[width=1cm, height=1cm, keepaspectratio]{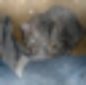}\\
Brightness & Contrast & Elastic & Pixelate & JPEG\\
\includegraphics[width=1cm, height=1cm, keepaspectratio]{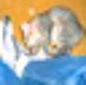}&
\includegraphics[width=1cm, height=1cm, keepaspectratio]{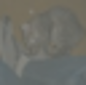}&
\includegraphics[width=1cm, height=1cm, keepaspectratio]{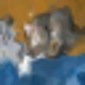}&
\includegraphics[width=1cm, height=1cm, keepaspectratio]{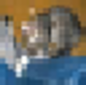}&
\includegraphics[width=1cm, height=1cm, keepaspectratio]{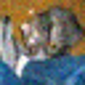}\\

\end{tabular}

    \caption{CIFAR-10-C \cite{hendrycks2018benchmarking} consists of 15 corrupted versions of the CIFAR-10 test dataset \cite{CIFAR10} with 5 levels of severity (level 5 here).}
    \label{CIFAR-10-C} 
\end{figure*}

\newpage
%
%
\bibliographystyle{splncs04}
%
\bibliography{references}

\end{document}